# A Multilingual Modeling Method for Span-Extraction Reading Comprehension

Gaochen Wu[1], Bin Xu[1], Lei Hou[1], Dejie Chang[2], Bangchang Liu[2]

[1]Computer Science and Technology, Tsinghua University, Beijing, China
[2]Beijing MoreHealth Technology Group Co. Ltd
[1]{wgc2019, xubin, houlei}@tsinghua.edu.cn
[2]{changdejie, liubangchang}@miao.cn

## ABSTRACT

Span-extraction reading comprehension models have made tremendous advances enabled by the availability of large-scale, high-quality training datasets. Despite such rapid progress and widespread application, extractive reading comprehension datasets in languages other than English remain scarce, and creating such a sufficient amount of training data for each language is costly and even impossible. An alternative to creating large-scale high-quality monolingual span-extraction training datasets is to develop multilingual modeling approaches and systems which can transfer to the target language without requiring training data in that language. In this paper, in order to solve the scarce availability of extractive reading comprehension training data in the target language, we propose a multilingual extractive reading comprehension approach called XLRC by simultaneously modeling the existing extractive reading comprehension training data in a multilingual environment using self-adaptive attention and multilingual attention. Specifically, we firstly construct multilingual parallel corpora by translating the existing extractive reading comprehension datasets (i.e., CMRC 2018) from the target language (i.e., Chinese) into different language families (i.e., English). Secondly, to enhance the final target representation, we adopt self-adaptive attention (SAA) to combine self-attention and inter-attention to extract the semantic relations from each pair of the target and source languages. Furthermore, we propose multilingual attention (MLA) to learn the rich knowledge from various language families. Experimental results show that our model outperforms the state-of-the-art baseline (i.e., RoBERTa_Large) on the CMRC 2018 task, which demonstrate the effectiveness of our proposed multi-lingual modeling approach and show the potentials in multilingual NLP tasks[1].

[1]The related code and model will be released in https://github.com/wgc20 soon.

## 1 Introduction

Reading Comprehension (RC) is a Question Answering (QA) task to evaluate how well computer systems understand human languages, where neural RC models have enjoyed the benefit of the availability of large-scale high-quality hand-annotated training datasets [1, 2]. As a large number of high-quality extractive reading comprehension datasets, such as SQuAD v1.1 [3], TriviaQA [4], NewsQA [5], SQuAD 2.0 [6] and Natural Questions [7], HotpotQA [8], CMRC [9], DRCD [10], DuReader [11] have been released, span-based reading comprehension has become an enormously popular topic in NLP.

Though tremendous advances in this field have been made by the NLP community in recent years, there are two challenges making training extractive reading comprehension systems in other languages more difficult: (1) Large-scale high-quality extractive reading comprehension datasets in languages other than English remain scarce. (2) Collecting such a sufficient amount of span-based reading comprehension training data for each language is costly and even impossible due to an existing large number of the world's languages with an astonishing breadth of linguistic phenomena. An alternative to creating large-scale high-quality monolingual span-extraction training datasets is to develop multilingual modeling approaches and systems which can transfer to the target language whether extractive reading comprehension training data is available or not in that language.

Asai et al. [12] proposed the span-extraction reading comprehension system for non-English languages with no extractive training datasets available, instead by an English extractive reading comprehension model and an attention-based neural machine translation (NMT) model. However, (1) the proposed system lacks semantic awareness between the passage and question sequences. Moreover, (2) attention-based answer alignment between different languages, especially for different language families with an astonishing breadth of linguistic phenomena used to express meaning, is different and could introduce additional noise. Cui et al. [13] proposed several back-translation approaches and Dual BERT for cross-lingual machine reading comprehension task. However, (1) back-translation approaches have to map the translated answer to the original answer span in the target language using Neural Machine Translation system (NMT). (2) Dual BERT only considers the <Passage, Question, Answer> in a bilingual context to learn the relations. While interesting attempts have been made, multilingual extractive reading comprehension modeling still has not been well-addressed.

In order to completely solve the scarce availability of extractive training data in target languages, we propose XLRC, a general multilingual extraction reading comprehension approach, by modeling the existing span-extraction reading comprehension training data in a multilingual environment. We aim to learn the rich hidden semantic knowledge from different language families and to make full use of all existing extractive training datasets in various languages.

To be specific, we firstly build multilingual parallel corpora by translating the existing extractive reading comprehension datasets (i.e., CMRC 2018) from the target language (i.e., Chinese) into different language families using the state-of-the-art Google Neural Machine (GNMT [14]) system, such as English and Japanese covered by TYDI QA [15] which chooses the 11 typologically diverse languages to represent top 100 human languages. Secondly, to enhance the final target representation, we adopt self-adaptive attention (SAA) to combine self-attention and inter-attention to extract the semantic relations from each pair of the target and source languages. Furthermore, we propose multilingual attention (MLA) to learn the rich general semantic knowledge from various language families. Multilingual attention allows our model to jointly attend to useful information from different representation subspaces in different language families.

It should also be noted that our proposed XLRC approach can generalize the learned knowledge to target languages whether extractive training data is available or not. This work focuses on the few-shot setting of extractive reading comprehension. To investigate the effectiveness of our proposed approach, we carry out several groups of preliminary experiments on two English extractive reading comprehension datasets: SQuAD v1.1 and SQuAD 2.0, and two Chinese span-based reading comprehension datasets: CMRC 2018 and DRCD. We choose CMRC 2018 as our target extractive reading comprehension task in this paper. Experimental results demonstrate the effectiveness of our proposed multi-lingual modeling approach by transferring the semantic knowledge learned from various existing datasets in different languages. In addition, we have already studied zero-shot extractive reading comprehension and achieved the state-of-the-art results on MLQA [16], we plan to share this zero-shot work in future.

Our contributions are summarized as follows: (1) we propose a multi-lingual extractive reading comprehension approach called XLRC by simultaneously modeling the existing extractive reading comprehension training data in a multilingual environment. (2) To learn the rich knowledge from various language families, we propose multilingual attention which allows the model to jointly attend to useful information from different representation subspaces of different language families. (3) Experimental results outperform the results of the state-of-the-art model (i.e., RoBERTa_Large), which demonstrate the effectiveness of our proposed multi-lingual modeling approach and show the potentials in multilingual NLP tasks.

## 2 Related Work

Question answering (QA) has been a popular research topic in NLP. There are a number of datasets available to tackle QA from various perspectives, such as span-extraction reading comprehension, cloze-style completion, and open domain QA [1, 3, 7], and so on. Large-scale extractive reading comprehension datasets such as SQuAD v1.1 [3], TriviaQA [4], NewsQA [5], SQuAD 2.0 [6] and Natural Questions [7] have seen attractive. Span-based reading comprehension becomes a dominant paradigm in QA and massive progress has been achieved on extractive reading comprehension by the community in recent years [17, 18]. Despite such popularity and widespread application, extractive QA datasets in languages other than English are comparatively rare. Moreover, collecting such large-scale high-quality annotated datasets is difficult and costly.

The increasing interest in cross-lingual extractive reading comprehension has been seen in recent years. Asai et al. [12] proposed the extractive reading comprehension system for non-English languages with no extractive training datasets available, instead by an English extractive reading comprehension model (i.e., BiDAF [19]) and an attention-based neural machine translation (NMT) model. However, the performance of the proposed system is quite low, and attention-based answer alignment between different languages with an astonishing breadth of linguistic phenomena used to express meaning is different and could introduce additional noise. Moreover, the proposed system lacks semantic awareness between the passage and question.

Cui et al. [13] proposed several back-translation approaches and Dual BERT for cross-lingual machine reading comprehension task. However, back-translation approaches have to align the translated answer with the source answer span in the original paragraph. The proposed Dual BERT can only be used for the condition when there is training data available in the target language. Moreover, Dual BERT only models the <Passage, Question, Answer> in a bilingual environment to learn the semantic relations. In a word, cross-lingual extractive reading comprehension modeling still has not been well-addressed.

In order to accelerate the research progress in the field of multilingual extractive reading comprehension, Lewis et al. [16] presented MLQA, a multi-way parallel extractive question answering evaluation benchmark in seven languages. Clark et al. [15] proposed TYDI QA −a question answering dataset covering 11 typologically diverse languages with 204K question-answer pairs. Artetxe et al. [20] created XQuAD, a dataset of 1190 SQuAD v1.1 samples from 240 paragraphs by manually translating the instances into ten languages. Their goals are to enable research progress toward building high-quality question answering systems in roughly the world top 100 languages and to encourage research on models that behave well across the linguistic phenomena and data scenarios of the world's languages.

To address the scarce availability of training data in target languages, we propose XLRC, a general multi-lingual extractive reading comprehension approach, to make full use of all existing extractive training datasets in various languages and to learn the rich hidden semantic knowledge from different language families by modeling the training data in a multilingual environment. Experimental results demonstrate the effectiveness of our proposed multilingual modeling approach by transferring the semantic knowledge learned from rich-resource languages and other existing training datasets to the target language.

## 3  Method

In order to take advantage of all existing extractive reading comprehension datasets in various languages and to further exploit the hidden semantic knowledge in different language families, we propose a multilingual modeling approach named XLRC, to simultaneously model the training data in a multilingual environment by using multilingual BERT [21] and multilingual attention based on multi-head attention mechanism [22]. The overall architecture of our proposed approach is shown in Figure 1.

### 3.1  Constructing Multilingual Parallel Corpora

In order to learn the semantic knowledge from various language families, we first need to construct multilingual parallel corpora by translating existing extractive reading comprehension datasets (i.e., CMRC 2018) in the target language (i.e., Chinese) to several representative languages. In our current research, we choose representative language families from the 11 typologically diverse languages covered by the information-seeking QA benchmark-TYDI QA which represents top 100 human languages.

In this paper, we firstly translate the Chinese extractive reading comprehension datasets (i.e., CMRC 2018 and DRCD) into English and Japanese, which represent two different language families, to form multilingual parallel corpora for multilingual modeling. It should also be noted that our multilingual modeling approach can support all the 11 typologically diverse languages in TYDI QA.

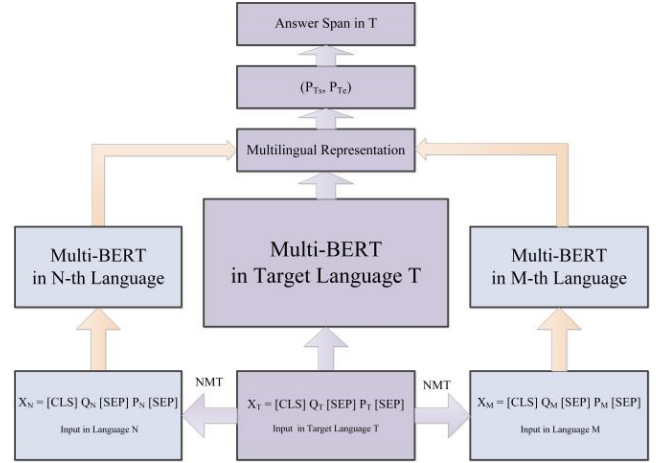

**Figure 1: The overview of our proposed multilingual modeling approach XLRC.**

Formally, given an extractive reading comprehension training sample (i.e., <Passage, Question, Answer>) in the target language T (i.e., Chinese) represented as <$P_T$, $Q_T$, $A_T$>, we represent each target language question-passage pair as an input token sequence $X_T$ for Multilingual BERT [21], shown as follows:

$$X_T = [CLS]Q_T[SEP]P_T[SEP] \qquad (1)$$

Similarly, we can obtain several translated multilingual parallel question-passage pairs of $X_T$ in different language families represented as input token sequences, such as $X_M$ in the language M and $X_N$ in the language N as follows:

$$X_M = [CLS]Q_M[SEP]P_M[SEP] \qquad (2)$$

$$X_N = [CLS]Q_N[SEP]P_N[SEP] \qquad (3)$$

### 3.2  Multilingual Modeling

In order to learn the general semantic information from various language families, we firstly use a shared multilingual BERT to encode the constructed multilingual parallel corpora to learn contextualized representations. Then we simultaneously model the extractive reading comprehension training data in a multilingual environment by using self-adaptive attention (SAA) and multilingual attention (MLA).

*3.2.1 Multilingual Encoder.* Bidirectional Encoder Representations from Transformers (BERT [21]), designed to learn deep bidirectional representations from unlabeled text by jointly conditioning on both left and right context in all layers, has shown astonishing performance in a broad range of diverse NLP tasks, which outperforms non-pretrained models by a large margin. As the multilingual BERT supports the top 100 languages with the largest Wikipedia, in this paper, we use multilingual BERT to encode the constructed multilingual parallel corpora.

To be specific, we use the input token sequences X (i.e., $X_T$, $X_M$ and $X_N$) for multilingual BERT to obtain contextualized representations B (i.e., $B_T$, $B_M$ and $B_N$, respectively) as follows:

$$B_T \in R^{L_T*h}, B_M \in R^{L_M*h}, B_N \in R^{L_N*h} \quad (4)$$

Where L indicates the length of the input token sequence and h represents the dimensionality of the hidden state (768 for multilingual BERT).

*3.2.2 Multilingual Modeling.* Typically, self-attention mechanism is used to relate different positions of a single sequence in order to calculate a more accurate representation of the sequence. Self-attention has been used successfully in a variety of NLP tasks including reading comprehension. In our multilingual modeling approach, to enrich the target language representation for final prediction, we firstly take advantage of self-attention to filter the irrelevant part of each representation (i.e., $B_T$, $B_M$ and $B_N$) in order to make the attention more accurate and then apply the softmax function to obtain self-attended representations as follows:

$$A_T = \text{softmax}(B_T \cdot B_T^T), A_T \in R^{L_T*h} \quad (5)$$

$$A_M = \text{softmax}(B_M \cdot B_M^T), A_M \in R^{L_M*h} \quad (6)$$

$$A_N = \text{softmax}(B_N \cdot B_N^T), A_N \in R^{L_N*h} \quad (7)$$

Then, we take advantage of multi-head inter-attention to further enhance the target representation by modeling the semantic relations between the passage and question in each pair of languages (i.e., the target language T and one of various language families, such as the language M or N). we calculate inter-attention of each pair of languages (i.e., T and M, T and N) as follows:

$$A_{TM} = B_T \cdot B_M^T, A_{TM} \in R^{L_T*L_M} \quad (8)$$

$$A_{TN} = B_T \cdot B_N^T, A_{TN} \in R^{L_T*L_N} \quad (9)$$

To utilize the benefit of both self-attention and inter-attention, we take advantage of Self-Adaptive Attention and calculate the dot product between self-adaptive attention and all source representations (i.e., $B_N$ and $B_M$) to obtain attended representations as follows:

$$\tilde{A}_{TM} = A_T \cdot A_{TM} \cdot A_M^T, \tilde{A}_{TM} \in R^{L_T*L_M} \quad (10)$$

$$\tilde{A}_{TN} = A_T \cdot A_{TN} \cdot A_N^T, \tilde{A}_{TN} \in R^{L_T*L_N} \quad (11)$$

$$C_M^{'} = \text{softmax}(\tilde{A}_{TM}) \cdot B_M, C_M^{'} \in R^{L_T*h} \quad (12)$$

$$C_N^{'} = \text{softmax}(\tilde{A}_{TN}) \cdot B_N, C_N^{'} \in R^{L_T*h} \quad (13)$$

Furthermore, we propose multilingual attention called MLA in order to learn the rich semantic knowledge from various language families by concatenating all attended representations in Equation (12) and (13). Multilingual attention allows the model to jointly attend to useful information from different representation subspaces of different language families.

$$Multilingual(T, M, N, ...) = Concat(C_M^{'}, C_N^{'}, ...) \quad (14)$$

In this paper, we only consider the target language T and two language families M and N form a multilingual environment, so the multilingual attention representation can be simplified as follows:

$$C^{'} = Multilingual(T, M, N) = Concat(C_M^{'}, C_N^{'}), \quad (15)$$
$$C^{'} \in R^{L_T*2h}$$

After obtaining multilingual attention representation, we use another fully connected layer along with residual layer normalization to get final multilingual-enhanced target representation as follows:

$$C = W_C C^{'} + b_C, W_C \in R^{2h*h}, C \in R^{L_T*h} \quad (16)$$

$$G_T = concat[B_T, LayerNorm(B_T + C)], \quad (17)$$
$$G_T \in R^{L_T*2h}$$

Finally, we use $G_T$ to perform final span prediction and calculate standard cross entropy loss for the answer span in the target language:

$$P_T^s = \text{softmax}\left(W_T^s G_T + b_T^s\right) \quad (18)$$

$$P_T^e = \text{softmax}\left(W_T^e G_T + b_T^e\right) \quad (19)$$

$$Loss_T = -\frac{1}{K}\sum_{k=1}^{K}\left(y_T^s \log(P_T^s) + y_T^e \log(P_T^e)\right) \quad (20)$$

Where W and b are trainable weights and bias parameters, superscript s and e for the start and end positions, T for the target language, respectively.

## 4 Experiments

### 4.1 Experimental Settings

We evaluate our proposed multilingual modeling method on two English extractive reading comprehension datasets, namely SQuAD v1.1 [3] and SQuAD 2.0 [6], and two public Chinese span-based reading comprehension datasets, such as CMRC 2018 [9] and DRCD [10].

Though there are some Chinese hand-annotated span-extraction training datasets available, but extractive reading comprehension in Chinese still has not been well-addressed. In this paper, we take advantage of existing extractive reading comprehension datasets (i.e., SQuAD v1.1, SQuAD 2.0, CMRC 2018 and DRCD) and choose CMRC 2018 as our target extractive reading comprehension task.

In this work, we choose English and Japanese from different language families and translate two Chinese extractive reading comprehension datasets, namely DRCD and CMRC 2018, into English and Japanese by using Google Neural Machine Translation (GNMT) system to construct multilingual parallel corpora in three languages (i.e., Chinese, English and Japanese). Furthermore, to further explore and exploit the knowledge in various existing extractive reading comprehension in different languages, we will choose to train our model on SQuAD v1.1 and SQuAD 2.0 for comparison in our experiments.

We modify the TensorFlow [23] version run_squad.py provided by BERT. All models in Table 1~5 are trained on a GeForce GTX 1080 Ti with 11 GB RAM. All models in Table 6 are trained on a GeForce Tesla V100 with 16 GB RAM. Due to the limitations of computational resources in our lab, we could not fully use the best hyperparameters for the datasets to explore the best performance of multilingual model method. We leave this work for our future research.

### 4.2 Baselines

As the multilingual BERT model supports the top 100 languages with the largest Wikipedia, we can take advantage of Multilingual BERT to build a simple Multilingual reading comprehension system. Therefore, we choose Multilingual BERT as one baseline in this work. we also choose the state-of-the -art model (i.e., RoBERTa_Large) over the CMRC 2018 task on the current official leaderboard as our another baseline for comparison in this paper.

In addition, we combine the key training parameters in our experiments as the form of (L, E, T, M), in which L represents learning rate and the value is L times of 1.0E-05, E denotes epochs, T and M are training batch size and max sequence length, respectively.

## 5 Results

### 5.1 Comparisons Between XLRC And Baselines

Table 1 shows the results for our proposed multilingual modeling XLRC model and various baselines on CMRC 2018. In this work, we could not use the best hyperparameters for CMRC 2018 due to the limitations of the computational resource in our lab, but this problem will be solved in our follow-up multilingual modeling research. According to Table 1, our proposed multilingual modeling model XLRC (#4) achieves the best F1 score of 82.55 over various baselines with the same hyperparameters (2, 2, 12, 128), demonstrating the effectiveness of our proposed multilingual modeling method. It should be noted that XLRC in Table 1 is not pre-trained on any existing extractive reading comprehension datasets other than CMRC 2018 training data, indicating that it is beneficial to improve the performance by just modeling the training data in a multilingual environment.

| Model | NO. | Learning Rate | Epochs | Batch Size | Max Sequence Length | F1 | EM |
|---|---|---|---|---|---|---|---|
| Multil-BERT | 1 | 2.0E-05 | 2 | 12 | 128 | 80.74 | 60.36 |
| CHN-BERT | 2 | 2.0E-05 | 2 | 12 | 128 | 81.24 | 61.35 |
| RoBERTa | 3 | 2.0E-05 | 2 | 12 | 128 | 82.37 | 63.47 |
| XLRC (w/o datasets) | 4 | 2.0E-05 | 2 | 12 | 128 | 82.55 | 62.13 |
| | 5 | 2.0E-05 | 2 | 12 | 160 | 84.37 | 63.78 |
| | 6 | 2.0E-05 | 2 | 12 | >160 | OOM | |

Table 1: Results on CMRC 2018 for multilingual modelling XLRC and various baselines. (w/o datasets) denotes that XLRC does not use any existing extractive reading comprehension datasets to enhance our proposed multilingual modelling model XLRC.

| Model | NO. | Hyperparameters | F1 | EM |
|---|---|---|---|---|
| Multi-BERT | 1 | (2,2,12,128) | 80.74 | 60.36 |
| XLRC (w/o datasets) | 2 | (2,2,12,128) | 82.55 | 62.13 |
| XLRC (w/ SQuAD v1.1) (2,2,12,128) | 3 | (2,2,12,128) | 84.05 | 62.94 |
| XLRC (w/ SQuAD 2.0) (2,2,12,128) | 4 | (2,2,12,128) | 83.97 | 63.59 |
| XLRC (w/ SQuAD v1.1) (2,2,12,300) | 5 | (2,2,12,128) | 83.92 | 62.85 |
| XLRC (w/ SQuAD v1.1) (2,2,12,128) | 6 | (2,2,12,160) | 85.00 | 64.90 |
| XLRC (w/ SQuAD 2.0) (2,2,12,128) | 7 | (2,2,12,160) | 85.21 | 64.65 |
| XLRC (w/ SQuAD v1.1) (2,2,12,300) | 8 | (2,2,12,160) | 85.34 | 65.08 |
| XLRC (w/ SQuAD 2.0) (2,2,12,300) | 9 | (2,2,12,160) | 85.78 | 65.18 |

Table 2: Results on the CMRC 2018 over multilingual modelling model XLRC pre-trained on SQuAD v1.1 or SQuAD 2.0 with different sets of the pretraining hyperparameters. XLRC (w/ SQuAD v1.1 with (2,2,12,128)) denotes that XLRC is pre-trained on SQuAD v1.1 with the hyperparameters (2, 2, 12, 128).

### 5.2 XLRC Pre-trained on SQuAD

Results on the CMRC 2018 task for XLRC pre-trained on SQuAD (including SQuAD v1.1 and SQuAD 2.0 separately) are shown in Table 2. Overall, all cases pre-trained on SQuAD significantly outperform the baselines (#1 and #2) over 1.5 points. Specifically,

XLRC with SQuAD v1.1 pre-trained weights could further improve the performance over XLRC without pre-training on SQuAD v1.1 (#2). Comparing between model 3 (#3) and model 4 (#5), we can observe that XLRC pre-trained with different sets of hypermeters on SQuAD v1.1 makes no difference on CMRC 2018 when the max sequence length is not enough during fine-tuning on CMRC 2018. We can also see that XLRC pre-trained on SQuAD 2.0 (#9) could obtains better F1 score than on SQuAD v1.1 (#8), suggesting that the more the training data is provided, the better the performance can be achieved.

| Model | NO. | Hyperparameters | F1 | EM |
|---|---|---|---|---|
| Multi-BERT | 1 | (2,2,12,128) | 80.74 | 60.36 |
| XLRC (w/o datasets) | 2 | (2,2,12,128) | 82.55 | 62.13 |
| XLRC (w/ Multi-DRCD) (2,2,12,160) | 3 | (2,2,12,160) | 85.04 | 64.52 |
| XLRC (w/ CHN-DRCD) (2,2,12,160) | 4 | (2,2,12,160) | 84.87 | 64.43 |
| XLRC (w/ Multi-DRCD) (2,2,6,256) | 5 | (2,2,12,160) | 85.56 | 65.92 |
| XLRC (w/ CHN-DRCD) (2,2,6,256) | 6 | (2,2,12,160) | 85.24 | 65.46 |
| XLRC (w/ Multi-DRCD) (2,2,6,256) | 7 | (2,2,2,256) | 86.40 | 65.45 |
| XLRC (w/ CHN-DRCD) (2,2,6,256) | 8 | (2,2,2,256) | 85.76 | 65.64 |

**Table 3: Results on CMRC 2018 over XLRC pre-trained on DRCD, including in only Chinese and in the multilingual environment (i.e., Chinese, English and Japanese). Multi-DRCD denotes DRCD is a multilingual dataset in Chinese, English and Japanese. CHN-DRCD denotes only in Chinese.**

### 5.3 XLRC Pre-trained on DRCD

Table 3 shows the results on the CMRC 2018 task over XLRC systems pretrained on DRCD with different training conditions. According to experimental results, we can observe that XLRC systems pretrained on multilingual DRCD training data can achieve better performance than on the DRCD data in only Chinese when using the same fine-tuning hyperparameters in the target task, for example, comparison between case 7 and case 8. Overall, all cases pretrained on DRCD datasets outperforms significantly the baselines ( #1 and #2 for this section).

### 5.4 XLRC Pre-trained on Multiple Datasets

Results on the CMRC 2018 task with XLRC systems pretrained on multiple existing high-quality span-based reading comprehension datasets are shown in Table 4. In this section, CMRC 2018 is our target task used for fine-tuning XLRC, and SQuAD v1.1, SQuAD 2.0 and DRCD (a multilingual parallel corpus in Chinese, English and Japanese), are our pretraining data used for training XLRC.

When pre-trained on a single dataset (#2, #3 and #4), XLRC can obtain better performance than the system without using pretrained weights (#1), demonstrating that learning the semantic knowledge from other existing datasets can improve the performance on the target downstream task. If we pretrain XLRC on multiple various datasets in a row (#5, #6, #7, #8), we observe that these pre-trained XLRC systems can further improve the performance on the target task, suggesting that XLRC can capture the knowledge from various datasets in various languages. Furthermore, the more datasets we adopt, the better performance we can achieve according to Table 4.

| NO. | XLRC Pretrained on Multiple Datasets | | | | F1 | EM |
|---|---|---|---|---|---|---|
| | 1 | 2 | 3 | 4 | | |
| 1 | CMRC (2,2,12,160) | | | | 84.37 | 63.78 |
| 2 | SQuAD v1.1 (2,2,12,300) | CMRC (2,2,12,160) | | | 85.34 | 65.08 |
| 3 | SQuAD 2.0 (2,2,12,300) | CMRC (2,2,12,160) | | | 85.78 | 65.18 |
| 4 | DRCD (2,2,6,256) | CMRC (2,2,12,160) | | | 85.56 | 65.92 |
| 5 | SQuAD v1.1 (2,2,12,300) | SQuAD 2.0 (2,2,12,300) | CMRC (2,2,12,160) | | 86.33 | 66.17 |
| 6 | SQuAD v1.1 (2,2,12,300) | DRCD (2,2,6,256) | CMRC (2,2,12,160) | | 86.47 | 67.32 |
| 7 | SQuAD 2.0 (2,2,12,300) | DRCD (2,2,6,256) | CMRC (2,2,12,160) | | 86.50 | 66.79 |
| 8 | SQuAD v1.1 (2,2,12,300) | SQuAD 2.0 (2,2,12,300) | DRCD (2,2,6,256) | CMRC (2,2,12,160) | 86.71 | 67.82 |

**Table 4: Results on CMRC 2018 over multilingual modelling model XLRC pretrained on various high-quality existing extractive reading comprehension datasets in various languages.**

It should also be noted that SQuAD v1.1 and SQuAD 2.0 are just English datasets, but DRCD for this section is a constructed multilingual dataset, indicating our proposed multilingual modeling model XLRC supports pretraining models in a multilingual parallel context.

### 5.5 Comparisons between XLRC and SOTA

In Table 5, we observe that our proposed model trained on SQuAD v1.1 could achieve better performance than the current state-of-the-art model with the same fine-tuning conditions (such as #1 and #2, #3 and #4), preliminarily demonstrating the effectiveness of our multilingual modelling method.

| Model | NO. | Hyperparameters | F1 | EM |
|---|---|---|---|---|
| RoBERTa_Large | 1 | (2,2,4,128) | 82.69 | 62.47 |
| XLRC (w/ {SQuAD v1.1 with (2,2,12,300)}) | 2 | (2,2,4,128) | 83.25 | 62.26 |
| RoBERTa_Large | 3 | (2,2,2,256) | 84.75 | 65.08 |
| XLRC (w/ {SQuAD v1.1 with (2,2,12,300)}) | 4 | (2,2,2,256) | 85.94 | 65.08 |
| XLRC (w/ {SQuAD v1.1 with (2,2,12,300)},{SQuAD2.0 with (2,2,12,300)}) | 5 | (2,2,2,256) | 86.61 | 66.12 |
| XLRC (w/ {SQuAD v1.1 with (2,2,12,300)},{SQuAD2.0 with (2,2,12,300)},{multi-DRCD with (2,2,6,256)}) | 6 | (2,2,2,256) | 86.90 | 66.82 |

**Table 5: Results on the CMRC 2018 task over our proposed model trained on SQuAD v1.1 with (2,2,12,300) and RoBERTa_Large, the current state-of-the-art model on the CMRC 2018 official leaderboard.**

| Model | NO. | Hyperparameters | F1 | EM |
|---|---|---|---|---|
| RoBERTa_Large | 1 | (2,2,4,256) | 84.75 | 65.08 |
| XLRC(our) (w/ {SQuAD v1.1 with (2,2,12,384)}, {SQuAD2.0 with (2,2,12,384)}, {multi-DRCD with (2,3,10,256)}) | 2 | (2,2,4,256) | 85.56 | 65.16 |
| RoBERTa_Large | 3 | (2,2,10,256) | OOM | OOM |
| XLRC(our) (w/ {SQuAD v1.1 with (2,2,12,384)}, {SQuAD2.0 with (2,2,12,384)}, {multi-DRCD with (2,3,10,256)}) | 4 | (2,2,10,256) | 86.15 | 65.18 |

**Table 6: Results on the CMRC 2018 task over our proposed model on multiple datasets and RoBERTa_Large. OOM means out of memory.**

Furthermore, in order to prove the effectiveness of multi-source learning for multilingual extractive reading comprehension task, we carry out several other experiments in Table 5 and Table 6, which are first trained on multiple datasets, such as experiment #6 in Table 5 trained on SQuAD v1.1 with (2,2,12,300), SQuAD 2.0 with (2,2,12,300) and multi-DRCD with (2,2,6,256) in a row, then train and fine-tune on CMRC 2018, finally perform answer prediction on the CMRC 2018 test data. The results show that the more sources we use for training, the better performance we can obtain.

In summary, our experimental results demonstrate that our proposed XLRC could learn the semantic knowledge by both pretraining on multiple various existing datasets in various languages and modeling the data in a multilingual environment, which are beneficial to improve the reading comprehension performance of the target tasks.

## 6 Conclusion

Despite tremendous advances have been made in the field of span-extraction reading comprehension in recent years, large-scale high-quality extractive QA datasets in languages other than English remain scarce, and collecting such a sufficient amount of training data for each language is costly, and even impossible, making training reading comprehension systems in other languages challenging. To make full use of all existing extractive training datasets in various languages and to learn the rich hidden semantic knowledge from different language families, we propose XLRC, a multilingual extractive reading comprehension method, to simultaneously model the existing extractive reading comprehension training data in a multilingual environment by using multilingual BERT and multilingual attention. Experimental results demonstrate the effectiveness of our proposed multilingual modelling by transferring the semantic knowledge learned from various existing datasets in different languages.

## ACKNOWLEDGMENTS

We would like to thank all anonymous reviewers for their thorough reviewing and providing constructive comments to improve our paper. This work was supported by Beijing MoreHealth Technology Group Co. Ltd. This work was also supported by the Ministry of Science and Technology via grant 2017YFB1401903 and 2018YFB1005101.